\documentclass[lettersize,journal]{IEEEtran}

\usepackage{amsmath,amsfonts}
\usepackage{array}
\usepackage[caption=false,font=normalsize,labelfont=sf,textfont=sf]{subfig}
\usepackage{textcomp}
\usepackage{stfloats}
\usepackage{url}
\usepackage{verbatim}
\usepackage{graphicx}
\usepackage{balance}
\usepackage{algorithm}
\usepackage{algpseudocode}
\usepackage{booktabs}
\usepackage{cite}
\usepackage{siunitx}
\usepackage{bm}

\def\BibTeX{{\rm B\kern-.05em{\sc i\kern-.025em b}\kern-.08em
    T\kern-.1667em\lower.7ex\hbox{E}\kern-.125emX}}

\begin{document}
\title{
Multimodal Adaptive Control for Safe Robotic Craniotomy Under Partial Observability}
\author{Xiao~Zhang$^{1\dagger}$, %
        Jiaxuan~Li$^{1\dagger}$, %
        Renzhen~Le, %
        Di~Wu, %
        Chao~Sun, %
        Jiachen~Zhu, %
        Haoyuan~Zhang, %
        Xiang~Li, %
        Jian~Liu, %
        Zhenzhi~Ying, %
        Pengfei~Zhang, %
        and~Liming~Shu$^*$,~\IEEEmembership{Member,~IEEE}
\thanks{This work has been submitted to the IEEE for possible publication. Copyright may be transferred without notice, after which this version may no longer be accessible.

$^\dagger$Equal contributions. $^*$Corresponding author. This work was supported by National Natural Science Foundation of China under Grant 52405456 and Grant W2621012. (\textit{corresponding author: Liming~Shu.})

Xiao~Zhang, Jiaxuan~Li, Di~Wu, Chao~Sun, Jiachen~Zhu, Haoyuan~Zhang, and Liming~Shu are with the Dalian University of Technology, Dalian 116024, China (e-mail: \protect\url{xz807113@gmail.com}; \protect\url{jiaxuan@mail.dlut.edu.cn}; \protect\url{diwu@mail.dlut.edu.cn}; \protect\url{sunchao@mail.dlut.edu.cn}; \protect\url{2563072882zjc@mail.dlut.edu.cn}; \protect\url{dlutecf@mail.dlut.edu.cn}; \protect\url{l.shu@dlut.edu.cn}).

Renzhen~Le is with the China Automotive Technology \& Research Center, Tianjin 300000, China (email: \protect\url{renzhenle0218@gmail.com};)

Xiang~Li and Jian~Liu are with the Second Hospital of Dalian Medical University, Dalian 116024, China (e-mail: 
\protect\url{lixiang_5007@163.com};
).

Zhenzhi~Ying and Pengfei~Zhang are with the Department of Mechanical Engineering, The University of Tokyo, Tokyo 113-8656, Japan (e-mail: 
\protect\url{ying@mfg.t.u-tokyo.ac.jp};
\protect\url{zhangpengfei@g.ecc.u-tokyo.ac.jp}).
}}

\markboth{
}%
{Zhang \MakeLowercase{\textit{et al.}}: RL-MACRO}

\maketitle

\begin{abstract}

Autonomous robotic craniotomy requires continuous regulation of tool-tissue interactions to mitigate mechanical overload and thermal damage while maintaining surgical efficiency. However, this process is inherently partially observable due to unknown, time-varying tissue properties and the inability to directly measure cutting temperatures under physical occlusion. To address these challenges, we propose RL-MACRO, a cybernetic closed-loop intelligence framework that couples multimodal perception, adaptive decision-making, and robotic execution. This framework empowers the surgical robot to autonomously perceive inaccessible states from partial sensory feedback and dynamically optimize its behaviors under uncertain environment. A CNN--LSTM observer first fuses force and sound feedback to reconstruct the hidden temperature state ($R^2=0.939$, $\mathrm{MAE}=1.717^\circ\mathrm{C}$). This reconstructed temperature, alongside multi-sensor features, forms the belief state for an offline Implicit Q-Learning (IQL) policy. A novel dual-head Actor dynamically coordinates the feed rate, spindle speed, and cutting depth to optimize efficiency within strict safety bounds. These decisions are seamlessly translated into spatial motions via online trajectory re-planning and velocity servoing. Experiments on bovine ribs and six \textit{ex vivo} goat skulls validate the system's robust perception, adaptive recovery from force/temperature excursions, and smooth execution on irregular surfaces, establishing a data-driven cybernetic paradigm for safe and efficient autonomous bone cutting.

\end{abstract}


\section{Introduction}

\IEEEPARstart{C}{raniotomy} is one of the most common and fundamental procedures in neurosurgery \cite{1}. With the increasing adoption of robot-assisted neurosurgery, high precision and stable manipulation have become achievable in cranial bone removal \cite{2}. However, autonomous craniotomy remains challenging because the interaction between a high-speed rotating tool and cranial tissue involves coupled mechanical loading, frictional heat generation, heterogeneous bone properties, and complex anatomical geometry. Excessive temperature may induce thermal necrosis and impair bone healing, while excessive cutting force may lead to bone fracture, tool damage, or sudden skull breakthrough, potentially injuring the dura mater \cite{3}. In addition, prolonged cutting duration may increase the risk of infection and other complications \cite{4}. Therefore, autonomous robotic craniotomy must simultaneously balance mechanical safety, thermal safety, and surgical efficiency.

This regulation problem is fundamentally limited by partial observability \cite{5}. Although sensors can be installed at the robot's end-effector or near the bones to monitor intraoperative states \cite{6}, the available information is often incomplete. Existing studies have investigated force, sound, vibration, and temperature sensing during robotic bone cutting. Jia \emph{et al.} \cite{7} fused acceleration and sound signals to identify tissue states during robotic laminectomy. Ying \emph{et al.} \cite{8} combined force modeling and acoustic features to recognize milling states and predict axial depth of cut. Temperature monitoring is typically performed using infrared thermal cameras or thermocouples \cite{9, 10, 11}. However, due to coolant flooding and the severe physical occlusion of the confined cutting region by the tool, high-frequency \textit{in vivo} temperature measurement using such equipment is highly intractable. Consequently, the robot receives only an incomplete information state of the cutting process, preventing controllers from accurately evaluating the consequences of previous actions. Although prior studies have mapped cutting parameters to bone temperature using theoretical models, support vector regression, or artificial neural networks \cite{12, 13, 14}, these methods mainly provide offline temperature estimation and do not reconstruct the latent thermal state from real-time feedback. As a result, the observability required for closed-loop thermomechanical regulation remains incomplete.

The subsequent challenge lies in converting sensory feedback into coordinated adaptive control. Bian \textit{et al.} \cite{15} proposed a model-free adaptive nonlinear force controller that regulates robotic motion to stabilize the normal contact force. Xia \textit{et al.} \cite{16} developed a fuzzy controller that adjusts the feed rate according to milling sound, while Sugita \textit{et al.} \cite{17} dynamically regulated the feed rate based on cutting-force deviations. Although these approaches successfully establish sensor-based feedback, most follow a single-input single-output (SISO) paradigm in which one process variable is adjusted to regulate a single principal physical response. Such formulations cannot effectively coordinate the feed rate, spindle speed, and cutting depth under coupled mechanical, thermal, and efficiency objectives. Moreover, the nonlinear and time-varying relationships between machining parameters and tissue responses make accurate analytical modeling exceedingly difficult. Reinforcement learning (RL) provides a compelling mechanism for learning coordinated state-to-action mappings directly from interaction data\cite{18}. Hathaway \textit{et al.} \cite{19} employed RL to adapt the feed rate, cutting depth, and compliance parameters during the robotic cutting of unknown materials. Li \textit{et al.} \cite{20} combined multisource tool-wear estimation with RL to reduce machining energy consumption. Deng \textit{et al.} \cite{21} developed an offline RL approach that learns control policies from historical factory data, thereby avoiding unsafe online exploration and reducing the dependence on high-fidelity simulators. These studies provide profound insights for parameter optimization and safe interaction in surgical robotics.

To address these challenges, we formulate autonomous craniotomy as a partially observable cybernetic regulation problem and propose RL-MACRO, a reinforcement-learning-based multimodal adaptive control framework. By seamlessly integrating latent-state observation, implicit state classification, offline RL decision-making, and dynamic spatial execution, RL-MACRO enables safe and efficient tool--tissue interaction. The main contributions of this paper are summarized as follows:

\begin{itemize}
\item[(1)] A comprehensive cybernetic closed-loop framework (perception--decision--execution) is established to generate adaptive, autonomous behaviors in highly uncertain cranial environments.

\item[(2)] A CNN--LSTM multimodal observer is developed to fuse force and sound feedback, successfully reconstructing the unmeasurable interfacial temperature essential for closed-loop thermomechanical regulation.

\item[(3)] An offline RL agent utilizing Implicit Q-Learning (IQL) with a novel dual-head Actor is designed. Guided by implicit state classification, it adapts the cutting depth with slowly varying cutting states while regulating the feed rate and spindle speed at high frequency to optimally balance safety and efficiency.

\item[(4)] A dynamic trajectory re-planning and velocity servoing mechanism is introduced to translate discrete policy decisions into kinematically smooth, spatially continuous robotic motions on complex cranial surfaces.

\item[(5)] Extensive \textit{ex vivo} validations on bovine ribs and six goat skulls demonstrate the system's robust temperature observability, proactive recovery from thermomechanical excursions, and excellent cross-specimen generalizability.
\end{itemize}

The remainder of this paper is organized as follows. Section II formulates the cybernetic problem. Section III introduces the experimental setup and offline dataset. Section IV details the RL-MACRO framework. Section V presents the experimental validations. Section VI discusses the translational significance and limitations. Section VII concludes the paper.

\section{Problem Formulation and Cybernetic Closed-Loop Architecture}

\begin{figure*}[t] 
    \centering
    \includegraphics[width=\textwidth]{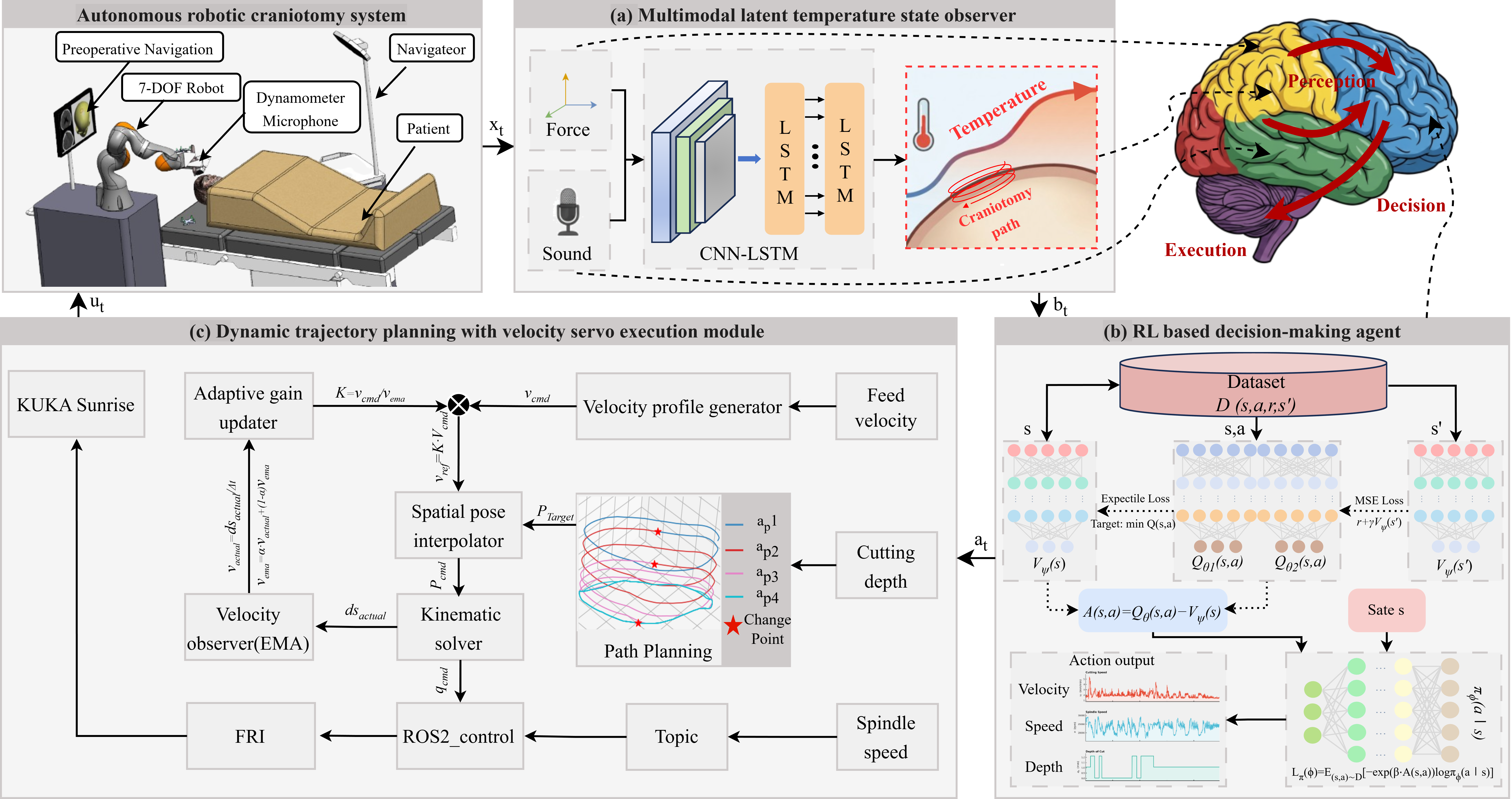}
    \caption{Cybernetic perception--decision--execution closed loop of RL-MACRO. (a) Multimodal latent temperature state observer. (b) RL-based decision-making agent. (c) Dynamic trajectory planning with velocity servo execution module.}
    \label{fig:system_workflow}
\end{figure*}

Consistent with our previous work \cite{22}, this study employs a spherical cutter for spiral bone-flap removal. Unlike traditional rigid machining, autonomous craniotomy requires the robot to continuously adapt to evolving tool--tissue thermomechanical reactions. Let $\mathbf{x}_t$ denote the latent physical state of the interaction at decision instant $t$ (e.g., local temperature, bone-layer characteristics, and material hardness). The interaction evolves as:
\begin{equation}
\mathbf{x}_{t+1} = f\left(\mathbf{x}_t, \mathbf{u}_t, \mathbf{w}_t\right)
\end{equation}
where $\mathbf{u}_t$ is the robotic control command, and $\mathbf{w}_t$ encapsulates unknown environmental disturbances (e.g., varying bone density, local thickness, and internal pores). Given the highly nonlinear and time-varying nature of $f(\cdot)$ and $\mathbf{w}_t$, deriving an accurate analytical dynamic model is intractable.

Due to severe surgical occlusion, the complete state $\mathbf{x}_t$ (particularly the safety-critical temperature) is unobservable. The robot relies solely on noisy sensory feedback, defined as $\mathbf{y}_t = g(\mathbf{x}_t) + \boldsymbol{\nu}_t$, where $\mathbf{y}_t = [F_t, S_t]^{\mathsf{T}}$ comprises the extracted force and sound features, and $\boldsymbol{\nu}_t$ is the measurement noise. To recover the inaccessible temperature state, a CNN--LSTM multimodal observer $\mathcal{E}_{\omega}$ reconstructs the temperature rise $\Delta\hat{T}_t$ from a historical sensory window $\mathbf{y}_{t-H:t}$. Concurrently, an implicit state classifier $C_{\phi}$ maps the immediate force--sound responses to a discrete cutting-regime probability $\mathbf{p}_t = C_{\phi}(F_t, S_t)$.

These components are concatenated to formulate an approximate belief state:
\begin{equation}
\hat{\mathbf{b}}_t = \left[ \Delta\hat{T}_t, F_t, S_t, \mathbf{p}_t \right]^{\mathsf{T}}
\end{equation}
Rather than attempting to reconstruct the full physical state $\mathbf{x}_t$, $\hat{\mathbf{b}}_t$ provides a compact, control-oriented representation containing the essential information requisite for thermomechanical regulation. Based on $\hat{\mathbf{b}}_t$, the RL agent generates the coordinated machining action $\mathbf{a}_t = \pi_{\theta}(\hat{\mathbf{b}}_t) = [v_{f, t}, n_t, a_{p, t}]^{\mathsf{T}}$, representing the target feed rate, spindle speed, and depth of cut, respectively. 

Subsequently, a trajectory re-planning and velocity servo module transforms the discrete action $\mathbf{a}_t$ into spatially continuous robotic joint commands:
\begin{equation}
\mathbf{u}_t = \Gamma\left(\mathbf{a}_t, \mathcal{G}, \mathbf{q}_t\right)
\end{equation}
where $\mathcal{G}$ represents the macroscopic skull geometry, and $\mathbf{q}_t$ denotes the current joint configuration.

As illustrated in Fig.~\ref{fig:system_workflow}, RL-MACRO establishes a comprehensive ``perception--decision--execution'' closed-loop framework. The execution of $\mathbf{u}_t$ drives the physical transition to $\mathbf{x}_{t+1}$, yielding new observations $\mathbf{y}_{t+1}$. This online cybernetic cascade can be elegantly summarized as:
\begin{equation}
\mathbf{x}_t \rightarrow \mathbf{y}_t \rightarrow \hat{\mathbf{b}}_t \rightarrow \mathbf{a}_t \rightarrow \mathbf{u}_t \rightarrow \mathbf{x}_{t+1}
\end{equation}
Through this recursive loop, the robot autonomously generates adaptive behaviors driven directly by the physical consequences of its preceding actions

\section{Experimental Platform and Offline Interaction Dataset}

\begin{figure*}[htbp] 
    \centering
    \includegraphics[width=\textwidth]{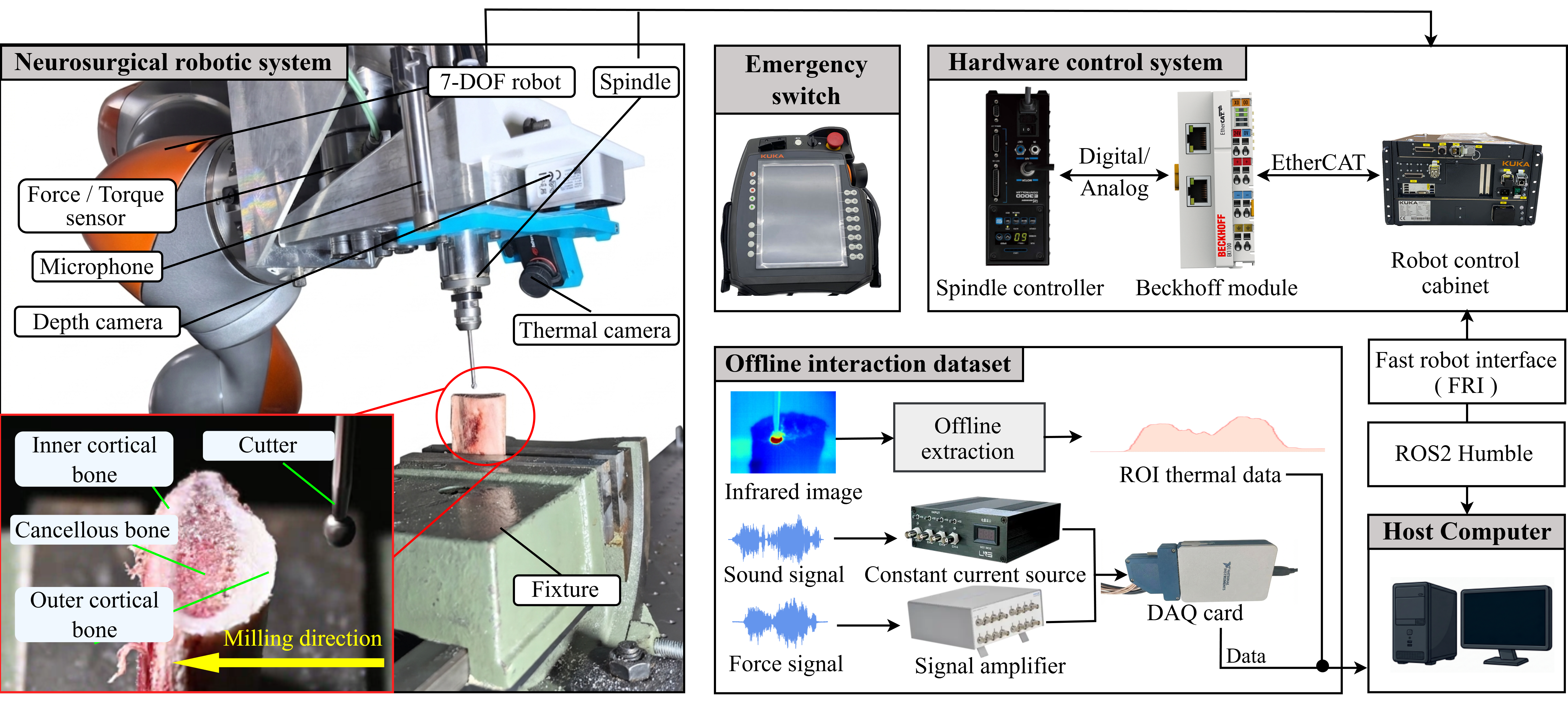}
    \caption{Experimental platform for synchronized multimodal interaction data acquisition, offline policy construction, and online closed-loop validation.}
    \label{fig:platform}
\end{figure*}

As depicted in Fig.~\ref{fig:platform}, the experimental platform comprises a 7-DoF robot (KUKA LBR iiwa 14) carrying a high-speed spindle (Nakanishi NR3060-AQC) and a $\SI{4}{\milli\metre}$ ball-end cutter. To capture multimodal interaction states, a microphone (BSWA MPA201), a 6-DoF dynamometer (Kistler 9306A), and an infrared (IR) camera (Infiray Xtherm T3Pro) are integrated at the end-effector. Hardware interfacing is managed via a Beckhoff I/O module for dynamic spindle regulation and an NI-9220 module for analog signal digitization.

Fresh bovine ribs---featuring human-like biomechanical properties \cite{13} and a ``cortical--cancellous--cortical'' architecture---were utilized for offline data collection to simulate craniotomy scenarios. During milling, the IR camera continuously tracked the newly exposed surface to establish ground-truth temperatures. A ROS 2-based host computer (Ubuntu 22.04) time-synchronized and logged all sensory and kinematic signals, constructing the foundational dataset required to train the multimodal observer and the offline RL agent.

\section{Cybernetic Closed-Loop Intelligence for Adaptive Robotic Craniotomy}

\subsection{Cybernetic Perception: Multimodal Latent Temperature State Observer}

\begin{figure}[htbp]
    \centering
    \includegraphics[width=\columnwidth]{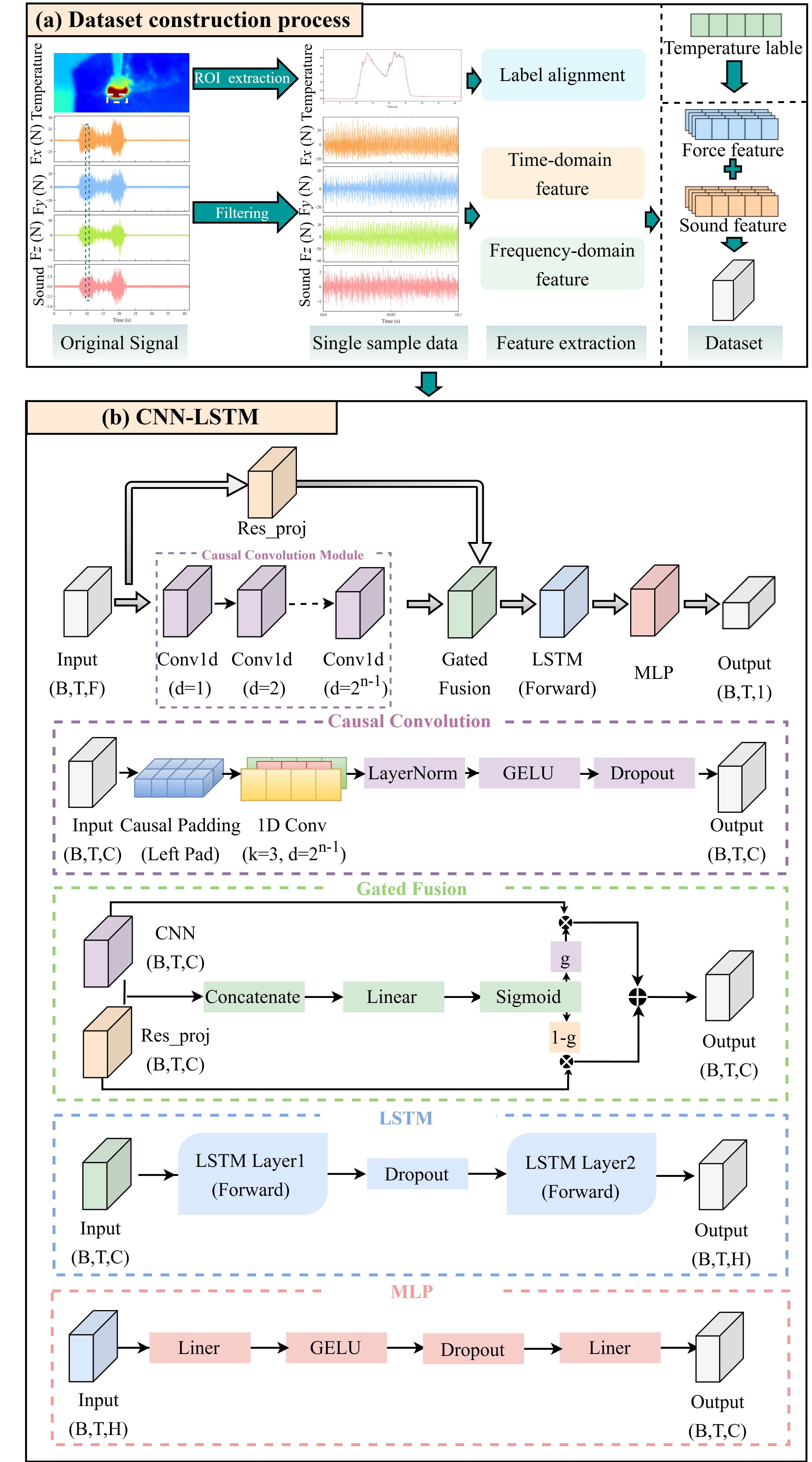}
    \caption{Causal multimodal temperature state observer for reconstructing the inaccessible cutting temperature rise from force and sound feedback.}
    \label{fig:CNN-LSTM}
\end{figure}

Because physical occlusion and continuous irrigation render direct temperature measurement intractable during realistic craniotomy, the perception layer must reconstruct this safety-critical latent state to enable closed-loop thermomechanical regulation \cite{24}. To ensure robust generalization, offline training data encompassing 60 distinct parameter combinations were collected: feed rates $v_f \in \{1, 2, 3, 4, 5\}\,\si{\milli\metre\per\second}$, spindle speeds $n \in \{15, 20, 25, 30\}\times 10^3\,\mathrm{rpm}$, and cutting depths $a_p \in \{0.5, 1.0, 1.5\}\,\si{\milli\metre}$. Here, force and sound histories serve as the observable inputs, while IR measurements provide the ground-truth temperature rise.

As illustrated in Fig.~\ref{fig:CNN-LSTM}, the proposed observer integrates a causal 1D-CNN, a gated residual fusion module, and a two-layer unidirectional LSTM. Given the synchronized force--sound input sequence $\mathbf{H}^{(0)}$, the $l$-th causal convolutional block is formulated as:
\begin{equation}
    \mathbf{H}^{(l)} = \mathrm{Dropout}\left( \mathrm{GELU}\left( \mathrm{LN}\left( \mathrm{Conv1D}(\mathbf{H}^{(l-1)}) \right) \right) \right)
\end{equation}
where $\mathrm{LN}(\cdot)$ denotes layer normalization. Strict left-padding ensures strict causality, preventing future information leakage, while dilation rates of 1 and 2 enlarge the receptive field.

To dynamically weigh the extracted convolutional features $\mathbf{H}_{c}$ against the linearly projected residual features $\mathbf{R}$, a learnable gating mechanism is introduced:
\begin{equation}
    \mathbf{G} = \sigma\left(\mathbf{W}_{g}[\mathbf{H}_{c};\mathbf{R}]+\mathbf{b}_{g}\right)
\end{equation}
\begin{equation}
    \mathbf{Z} = \mathbf{G}\odot \mathbf{H}_{c} + (1-\mathbf{G})\odot \mathbf{R}
\end{equation}
where $\sigma(\cdot)$ is the Sigmoid function and $\odot$ denotes element-wise multiplication. This fusion preserves original sensory criticalities while emphasizing nonlinear convolutional features when interaction conditions dictate.

The fused sequence $\mathbf{Z}$ is subsequently processed by the LSTM to capture long-term thermal inertia. A regression head ultimately maps the hidden state to the reconstructed temperature rise $\Delta\hat{T}_t = \mathcal{E}_{\omega}(\mathbf{y}_{t-H:t})$.

\begin{figure}[htbp]
    \centering
    \includegraphics[width=\columnwidth]{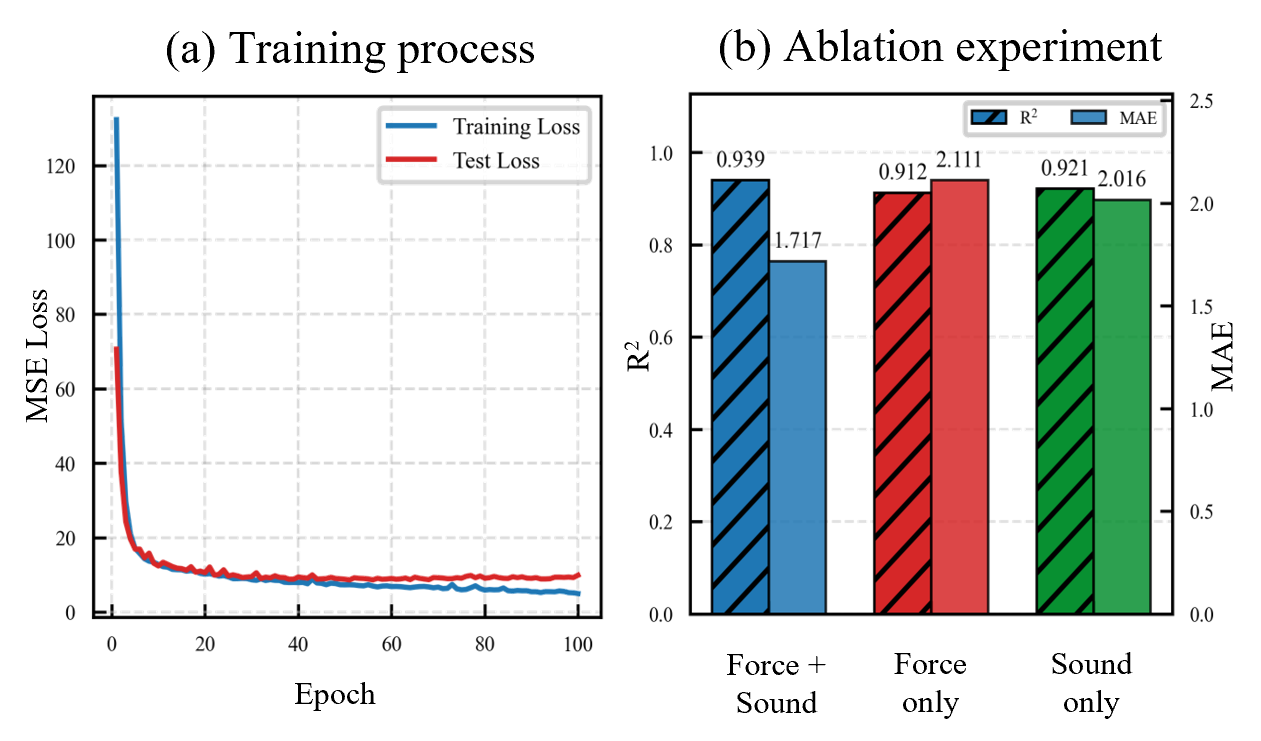}
    \caption{Training and ablation results of the multimodal thermal-state observer. (a) Training and validation loss curves. (b) Comparison between multimodal and single-modality temperature state perception.}
    \label{fig:CNN-LSTM result}
\end{figure}

As depicted in Fig.~\ref{fig:CNN-LSTM result}(a), the observer converged stably. Ablation studies (Fig.~\ref{fig:CNN-LSTM result}(b)) confirm that fusing force and sound provides vastly superior perception over single-modality baselines. Ultimately, the multimodal observer achieved an $R^2$ of $0.939$ and a Mean Absolute Error (MAE) of $\SI{1.717}{\celsius}$ on the offline test set, robustly securing high-fidelity thermal feedback for the subsequent RL agent.

\subsection{Cybernetic Decision-Making: Safety-Aware Regulation via Offline RL}

\begin{figure*}[htbp]
    \centering
    \includegraphics[width=\textwidth]{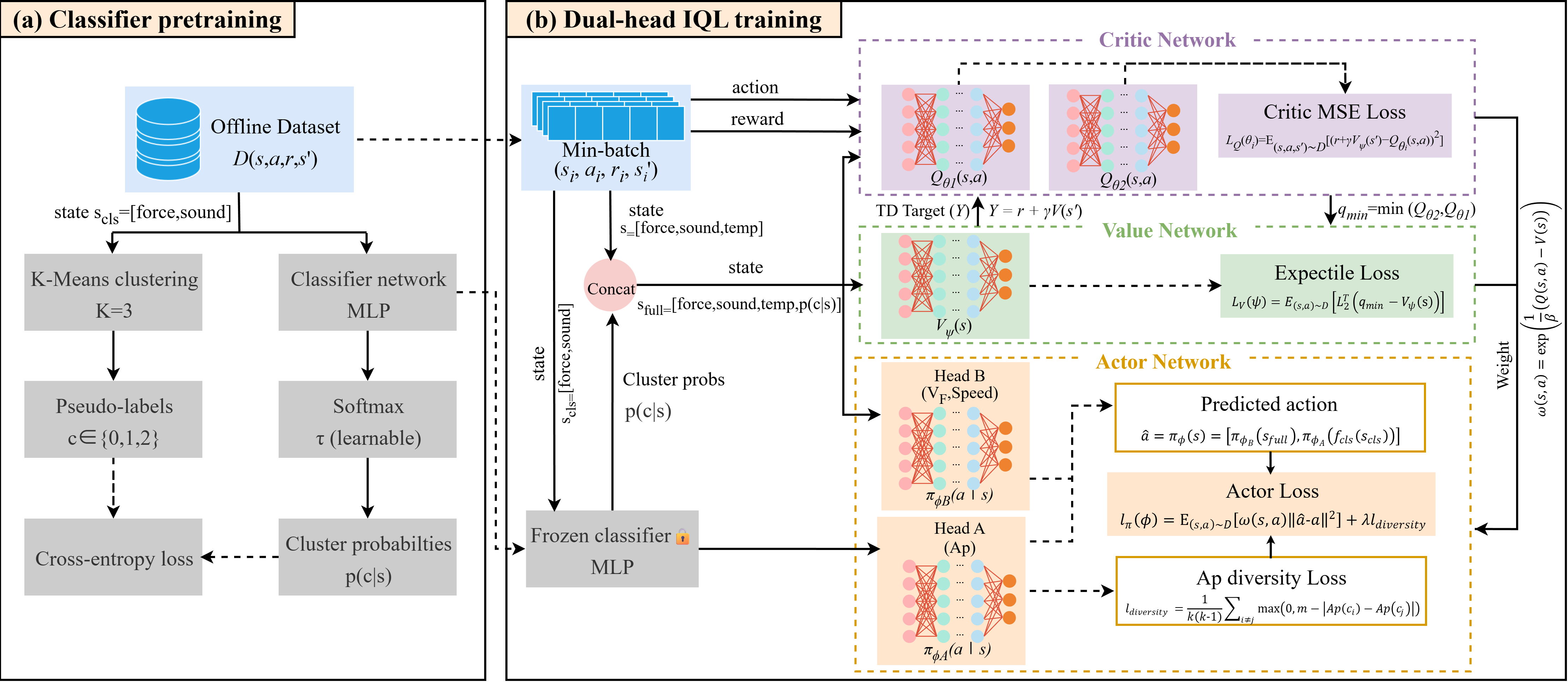}
    \caption{Implicit state classifier-guided dual-head IQL framework. The multimodal observer and implicit state classifier construct an approximate belief state, from which the structured Actor generates coordinated feed-rate, spindle-speed, and cutting-depth actions.}
    \label{fig:iql}
\end{figure*}

To convert the perceived latent state into coordinated adaptive behavior, we formulate the non-linear, strongly coupled thermomechanical regulation problem as a partially observable offline RL task \cite{25}. The offline interaction dataset is defined as $\mathcal{D} = \{(\mathbf{o}_t,\mathbf{a}_t,r_t,\mathbf{o}_{t+1})\}_{t=1}^{N}$. The control-oriented observation merges the reconstructed temperature with measurable mechanics: $\mathbf{o}_t = [\Delta\hat{T}_t, F^{\mathrm{rms}}_t, S^{\mathrm{ent}}_t]^{\mathsf{T}}$. The action vector encompasses the three continuous machining parameters: $\mathbf{a}_t = [v_{f,t}, n_t, a_{p,t}]^{\mathsf{T}}$, representing the feed rate, spindle speed, and cutting depth, respectively.

The reward function balances material-removal efficiency with safety constraints. Efficiency is evaluated via the Material Removal Rate (MRR), derived from the cutter's geometry:
\begin{equation}
    \mathrm{MRR}_t = \frac{1}{2}R^2(\theta_t-\sin\theta_t) \cdot v_{f,t}, \quad \theta_t = 2\arccos\left(\frac{R-a_{p,t}}{R}\right)
\end{equation}
where $R$ is the cutter radius and $\theta_t$ is the contact angle, yielding the efficiency reward $r_{\mathrm{eff},t} = w_{\mathrm{eff}} \cdot \mathrm{MRR}_t$. To suppress thermomechanical overload, a quadratic ReLU penalty activates when safety thresholds ($F_{\max}$, $\Delta T_{\max}$) are breached:
\begin{equation}
    \mathcal{P}_{\mathrm{safe},t} = w_{\mathrm{safe}} \left[ \left( F^{\mathrm{rms}}_t-F_{\max} \right)_{+}^{2} + \left( \Delta\hat{T}_t-\Delta T_{\max} \right)_{+}^{2} \right]
\end{equation}
where $(x)_{+}=\max(0,x)$ and $w_{\mathrm{safe}}$ is the penalty weight. The overall immediate reward is thus $r_t = r_{\mathrm{eff},t} - \mathcal{P}_{\mathrm{safe},t}$.

\begin{figure}[htbp]
    \centering
    \includegraphics[width=\columnwidth]{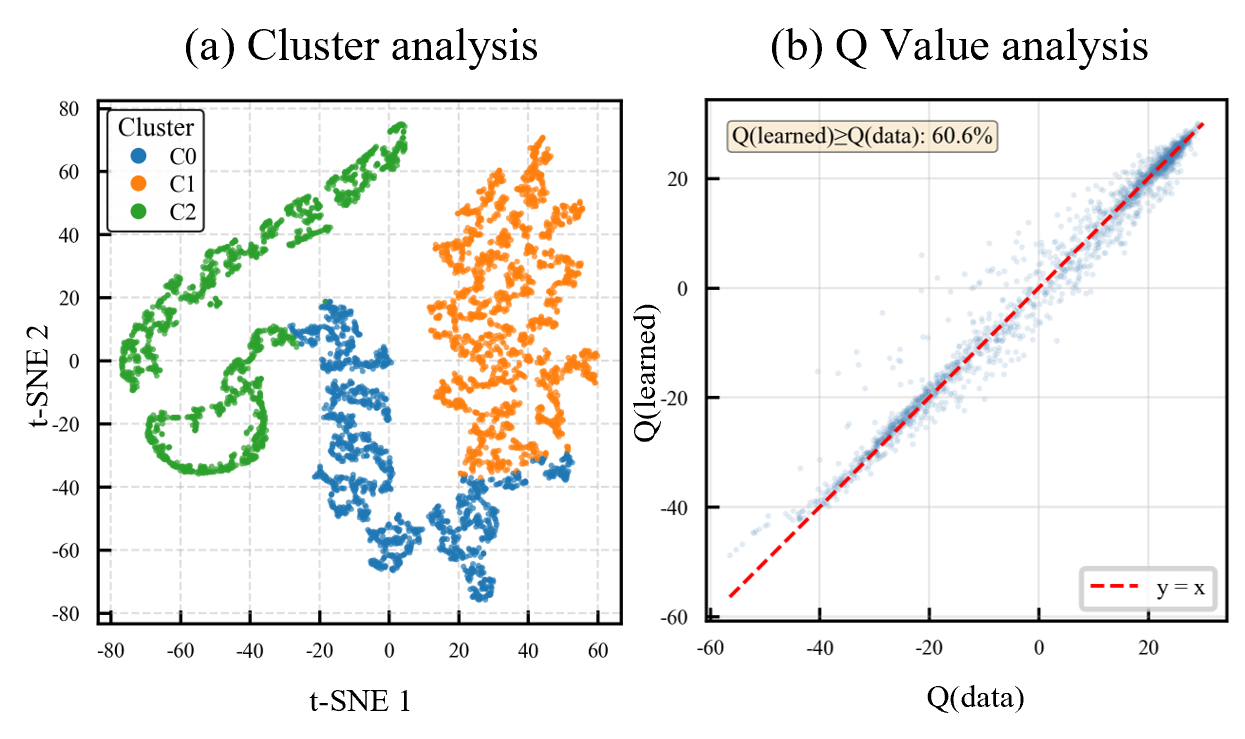}
    \caption{Clustering of implicit cutting state and offline policy evaluation. (a) Force--sound clustering of latent states. (b) Comparison between the estimated Q-values of the learned policy and the behavior actions.}
    \label{fig:iql result}
\end{figure}

To explicitly distinguish implicit cutting regimes, K-means clustering is initially applied to the mechanical features $\mathbf{o}^{c}_t =[F^{\mathrm{rms}}_t,S^{\mathrm{ent}}_t]$. The resulting cluster indices (Fig.~\ref{fig:iql result}(a)) serve as pseudo-labels to supervise an MLP classifier $C_{\phi}$. Post-training, the frozen classifier outputs a state probability vector $\mathbf{p}_t = C_{\phi}(\mathbf{o}^{c}_t)$, which is concatenated with the observation to form the approximate belief state $\hat{\mathbf{b}}_t = [\mathbf{o}_t;\mathbf{p}_t]$. This ensures the policy conditions on both transient sensory deviations and macroscopic tissue variations.

A structured dual-head Actor is designed to decouple the operational frequencies of the action variables. The first branch relies solely on $\mathbf{p}_t$ to predict the cutting depth ($\hat{a}^{p}_t = \tanh(h_{p}(\mathbf{p}_t))$), enforcing consistency within the same cutting regime and preventing mechanical chattering. The second branch receives the full belief state $\hat{\mathbf{b}}_t$ to continuously fine-tune the feed rate and spindle speed at high frequency: $[\hat{a}^{v}_t,\hat{a}^{n}_t] = \tanh(h_{vn}(\hat{\mathbf{b}}_t))$. The final policy output is concatenated as:
\begin{equation}
    \pi_{\theta} (\hat{\mathbf{b}}_t) = \left[ \hat{a}^{v}_t, \hat{a}^{n}_t, \hat{a}^{p}_t \right]^{\mathsf{T}}
\end{equation}

Following the Implicit Q-Learning (IQL) paradigm \cite{26}, the Value network $V_{\eta}$ is updated via expectile regression, while two independent Critics ($Q_{\psi_1}, Q_{\psi_2}$) minimize the Bellman error. Crucially, all value networks operate on the extended belief state $\hat{\mathbf{b}}_t$. The Actor is updated via Advantage-Weighted Behavioral Cloning with exponential weights $w_t$. To prevent mode collapse in the decoupled architecture---where depth outputs for different clusters might erroneously converge---a novel diversity regularization term $\mathcal{L}_{\mathrm{div}}$ is integrated:
\begin{equation}
    \mathcal{L}_{\pi}(\theta) = \mathbb{E}_{\mathcal{D}} \left[ w_t \left\| \pi_{\theta}(\hat{\mathbf{b}}_t)-\mathbf{a}_t \right\|_2^2 \right] + \lambda_{\mathrm{ap}}\mathcal{L}_{\mathrm{div}}
\end{equation}
\begin{equation}
    \mathcal{L}_{\mathrm{div}} = \frac{1}{|\mathcal{C}|} \sum_{i<j} \max \left( 0, m - \left| h_p(\mathbf{e}_i)-h_p(\mathbf{e}_j) \right| \right)
\end{equation}
where $\mathbf{e}_i$ and $\mathbf{e}_j$ represent the one-hot encoded vectors of different clusters, $m$ is the desired minimum depth interval margin, and $\lambda_{\mathrm{ap}}$ is the regularization weight.

Post-training evaluations (Fig.~\ref{fig:iql result}(b)) reveal that the learned policy yields a higher estimated Q-value than the behavioral baseline in $60.6\%$ of the offline states. This confirms that the customized IQL policy successfully extracts and amalgamates higher-value safety--efficiency behaviors from the suboptimal dataset, establishing a robust foundation for online closed-loop execution.

\begin{algorithm}[t]
\caption{Implicit State Classifier-Guided Dual-Head IQL}
\label{alg:dual_head_iql}
\small
\begin{algorithmic}[1]

\Require Offline dataset $\mathcal{D}$, number of clusters $K$, maximum iterations $M$, learning rates $\alpha_V$, $\alpha_Q$, and $\alpha_{\pi}$
\Ensure Optimized Actor policy $\pi_{\theta}$

\State \textbf{/* Phase 1: Implicit Cutting-State Construction */}
\State Extract force--sound features $\mathcal{O}^{c} = \{ (F^{\mathrm{rms}}_t,S^{\mathrm{ent}}_t) \}_{t=1}^{N}$ from $\mathcal{D}$
\State $\mathcal{L}_{\mathrm{pseudo}} \leftarrow \Call{KMeans}{\mathcal{O}^{c},K}$
\State Train classifier $C_{\phi}:\mathcal{O}^{c}\rightarrow\mathcal{L}_{\mathrm{pseudo}}$
\State Freeze classifier parameters $\phi$

\State \textbf{/* Phase 2: Offline Policy Learning */}
\State Initialize $\pi_{\theta}$, $Q_{\psi_1}$, $Q_{\psi_2}$, and $V_{\eta}$

\For{$m=1,2,\ldots,M$}

    \State Sample $(\mathbf{o}_t,\mathbf{a}_t,r_t,\mathbf{o}_{t+1},d_t) \sim\mathcal{D}$
    \State $\mathbf{p}_t \leftarrow C_{\phi}(\mathbf{o}^{c}_t)$, $\quad \mathbf{p}_{t+1} \leftarrow C_{\phi}(\mathbf{o}^{c}_{t+1})$
    \State $\hat{\mathbf{b}}_t \leftarrow [\mathbf{o}_t;\mathbf{p}_t]$, $\quad \hat{\mathbf{b}}_{t+1} \leftarrow [\mathbf{o}_{t+1};\mathbf{p}_{t+1}]$

    \State \textit{// Update Value Network}
    \State $\eta \leftarrow \eta - \alpha_V \nabla_{\eta} \mathcal{L}_{V}(\eta)$

    \State \textit{// Update Critic Networks}
    \State $y_t \leftarrow r_t + \gamma(1-d_t) V_{\eta}(\hat{\mathbf{b}}_{t+1})$
    \For{$i\in\{1,2\}$}
        \State $\psi_i \leftarrow \psi_i - \alpha_Q \nabla_{\psi_i} \mathcal{L}_{Q}(\psi_i)$
    \EndFor

    \State \textit{// Update Structured Actor}
    \State $A_t \leftarrow \min_{i\in\{1,2\}} Q_{\psi_i}(\hat{\mathbf{b}}_t,\mathbf{a}_t) - V_{\eta}(\hat{\mathbf{b}}_t)$
    \State $w_t \leftarrow \operatorname{clip} \left( \exp(A_t/\beta), 0, w_{\max} \right)$
    \State $\mathcal{L}_{\mathrm{Actor}} \leftarrow \mathcal{L}_{\mathrm{AWBC}} + \lambda_{\mathrm{ap}} \mathcal{L}_{\mathrm{div}}$
    \State $\theta \leftarrow \theta - \alpha_{\pi} \nabla_{\theta} \mathcal{L}_{\mathrm{Actor}}$

\EndFor

\State \Return $\pi_{\theta}$

\end{algorithmic}
\end{algorithm}

\subsection{Cybernetic Execution: Spatially Continuous Action Mapping and Servo Feedback}

\begin{figure*}[htbp]
    \centering
    \includegraphics[width=\textwidth]{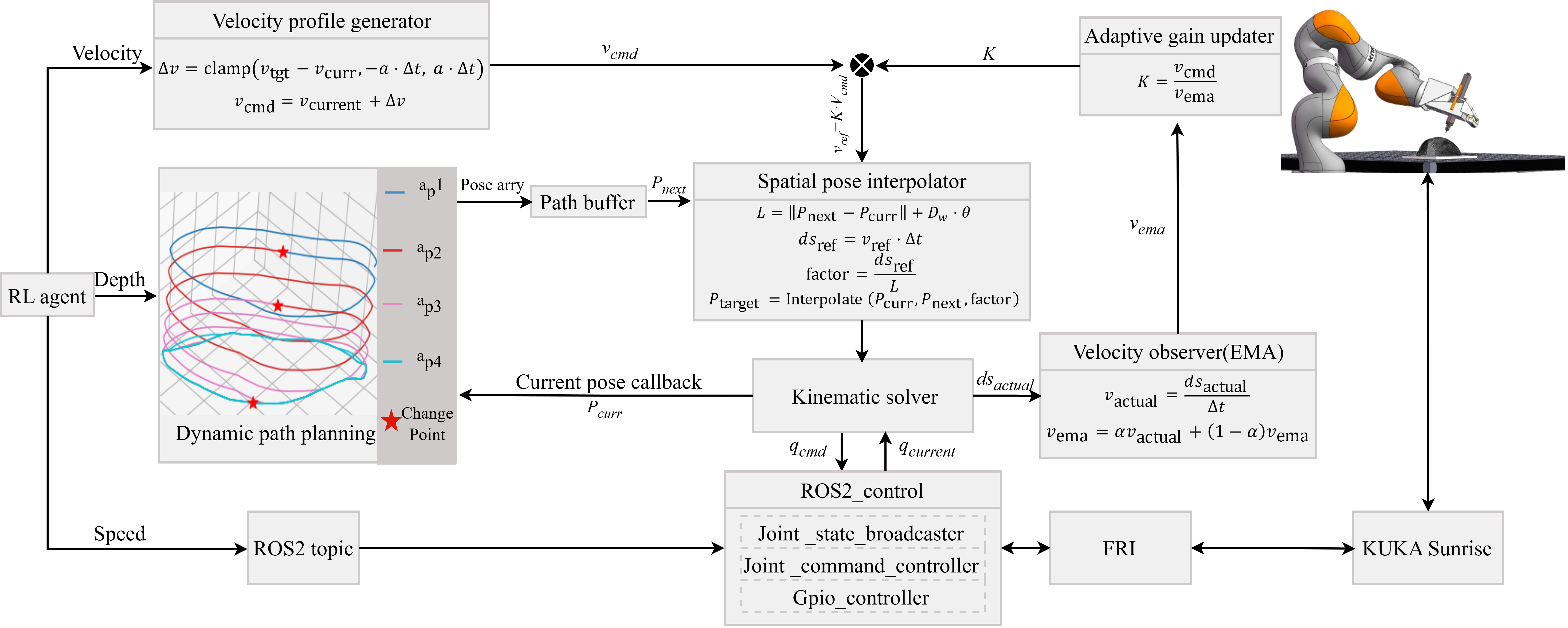}
    \caption{Architecture of dynamic trajectory re-planning and velocity servo for continuity-preserving action realization.}
    \label{fig:control}
\end{figure*}

Because the RL agent generates discrete, high-level machining actions $\mathbf{a}_t=[v_{f,t}, n_t, a_{p,t}]$, a trajectory re-planning and velocity servo layer is required to translate these decisions into kinematically feasible, continuous robot motions. This module serves as the physical execution interface of the cybernetic loop (Fig.~\ref{fig:control}). Building upon our prior contour-extraction framework \cite{22}, we introduce a local adaptive blending method to support online pitch updates and residual trajectory re-planning.

Let $\Delta H(\phi)$ denote the local axial clearance between the inner and outer skull contours at angular phase $\phi$. Given an initial pitch $P$ (representing depth of cut), the nominal cumulative axial depth is $D(\phi) = \frac{P}{2\pi}(\phi-\phi_{\mathrm{start}})$. To dynamically conform to heterogeneous cranial geometry, a local blending coefficient is defined:
\begin{equation}
\alpha(\phi) = \operatorname{sat}_{[0,1]} \left[ \frac{D(\phi)} {\max\left(\Delta H(\phi),\epsilon\right)} \right]
\label{eq:local_blending}
\end{equation}
The spatial trajectory $\bm{\xi}(\phi) = [\rho(\phi), z(\phi)]^\top$ is subsequently interpolated between the contours:
\begin{equation}
\bm{\xi}(\phi) = \left[1-\alpha(\phi)\right] \bm{\xi}_{\mathrm{outer}}\left(\tilde{\phi}_{\mathrm{outer}}\right) + \alpha(\phi) \bm{\xi}_{\mathrm{inner}}\left(\tilde{\phi}_{\mathrm{inner}}\right)
\label{eq:trajectory_interpolation}
\end{equation}
This adaptive blending strictly enforces a constant axial cut depth $P$ per revolution before reaching the inner boundary ($\alpha < 1$).

Crucially, when the RL agent updates the pitch to $P_j$ at event $j$, the physically accumulated depth $D_j$ is measured from the current Cartesian pose $\mathbf{x}_j$. The remaining trajectory's depth function is instantly reconstructed as:
\begin{equation}
D(\phi) = D_j + \frac{P_j}{2\pi} \left( \phi-\phi_j \right), \qquad \phi\geq\phi_j
\label{eq:online_pitch}
\end{equation}
Substituting Eq.~\eqref{eq:online_pitch} back into Eq.~\eqref{eq:local_blending} re-parameterizes the unexecuted trajectory. This mechanism generates a piecewise constant-pitch spiral that mathematically guarantees $C^0$ spatial continuity at $\mathbf{x}_j$, while quaternion hemisphere consistency concurrently prevents rotational jumps.

To smoothly track this dynamic trajectory, a velocity servo strategy utilizing composite arc-length interpolation is implemented. First, to isolate operating system (OS) scheduling jitter, the control period is clamped to $\Delta t_k \in [\Delta t_{\min}, \Delta t_{\max}]$. The RL-commanded feed rate $v_{\mathrm{ref}}$ is then smoothed via an acceleration-limited first-order filter to prevent mechanical shocks:
\begin{equation}
    v_k = v_{k-1} + \operatorname{clip}\left(v_{\mathrm{ref}}-v_{k-1}, -a_{\max}\Delta t_k, a_{\max}\Delta t_k\right)
    \label{eq:speed_smoothing}
\end{equation}
where $a_{\max}$ is the maximum allowable Cartesian acceleration.

Let $\bm{X}_k=\{\bm{p}_k,\bm{R}_k\}$ and $\bm{X}_{k+1}$ denote the current and next discrete target poses. A composite Cartesian arc length $d_k$ synchronously evaluates translational and rotational distances:
\begin{equation}
    d_k = \sqrt{\left\|\bm{p}_{k+1}-\bm{p}_k\right\|^2 + \left(w_R\theta_k\right)^2}
    \label{eq:composite_arc}
\end{equation}
where $\theta_k$ is the relative rotation angle derived from $\bm{R}_{k+1}\bm{R}_k^{\mathsf{T}}$, and $w_R$ maps orientation to translational metrics. Given the nominal advancing step $\Delta s_k = K_k v_k\Delta t_k$, the interpolation ratio becomes:
\begin{equation}
    \eta_k = \min\left(\frac{\Delta s_k}{\max(d_k,\epsilon)}, 1\right)
    \label{eq:interpolation_ratio}
\end{equation}
If $\eta_k = 1$, the trajectory advances to the next waypoint; otherwise, incremental interpolation is executed, and expected joint angles $\bm{q}_k$ are resolved via inverse kinematics.

Finally, to compensate for spatial discretization and cycle fluctuations, an Exponential Moving Average (EMA) observer tracks the physically accumulated arc length $\Delta s_k^{\mathrm{acc}}$:
\begin{equation}
    \bar v_k = \lambda_v \frac{\Delta s_k^{\mathrm{acc}}}{\Delta t_k} + \left(1-\lambda_v\right)\bar v_{k-1}
    \label{eq:speed_ema}
\end{equation}
An adaptive gain law then dynamically corrects the step-length multiplier $K_k$:
\begin{equation}
    K_k = \lambda_K \operatorname{clip}\left[\frac{v_k}{\max(\bar v_k,\epsilon)}, K_{\min}, K_{\max}\right] + \left(1-\lambda_K\right)K_{k-1}
    \label{eq:adaptive_gain}
\end{equation}

This comprehensive execution interface robustly transforms discrete policy actions into fluid, $C^0$-continuous spatial motions, completely detailed in Algorithm~\ref{alg:control}.

\begin{algorithm}[t]
\caption{Online Dynamic Trajectory Replanning and Adaptive Velocity Servo}
\label{alg:control}
\small
\begin{algorithmic}[1]
\Require Initial pitch $P_0$, reference speed $v_{\mathrm{ref}}$, trajectory buffer $\mathcal{B}$
\Ensure Joint command $\mathbf{q}_k^d$

\State $\mathcal{B}\leftarrow \Call{PlanInitialTrajectory}{P_0}$,
$v_0\leftarrow0$, $\bar v_0\leftarrow0$, $K_0\leftarrow1$

\While{robot servo is active}

    \If{new pitch command $P_j$ is received}
        \State $\{\mathbf{x}_j,\mathbf{q}_j\}\leftarrow \Call{ReadCurrentState}{}$
        \State $D_j\leftarrow z_j-z_{\mathrm{outer}}(\tilde{\phi}_{\mathrm{outer},j})$
        \If{$D_j<\Delta H_{\max}$}
            \State $D(\phi)\leftarrow D_j+\dfrac{P_j}{2\pi}(\phi-\phi_j)$, $\phi\ge\phi_j$
            \State $\mathcal{B}\leftarrow \Call{ReplanTrajectory}{\mathbf{x}_j,\mathbf{q}_j,D(\phi)}$
        \Else
            \State $\Call{RejectCommand}{P_j}$
        \EndIf
    \EndIf

    \State $\Delta t_k\leftarrow
    \operatorname{clip}(\Delta t_k^{\mathrm{meas}},\Delta t_{\min},\Delta t_{\max})$

    \State $v_k\leftarrow v_{k-1}
    +\operatorname{clip}(v_{\mathrm{ref}}-v_{k-1},
    -a_{\max}\Delta t_k,a_{\max}\Delta t_k)$

    \State $\Delta s_k^{d}\leftarrow K_{k-1}v_k\Delta t_k$
    \State $\mathbf{X}_k^\star\leftarrow \Call{Interpolate}{\mathcal{B},\Delta s_k^{d}}$
    \State $\mathbf{q}_k^d\leftarrow \Call{InverseKinematics}{\mathbf{X}_k^\star,\mathbf{q}_{k-1}}$

    \If{$\Call{CheckSafetyConstraints}{\mathbf{q}_k^d}$}
        \State $\Call{SendToRobot}{\mathbf{q}_k^d}$
        \State $\Delta s_k^{a}\leftarrow \Call{MeasureActualArcLength}{}$
        \State $\bar v_k\leftarrow
        \lambda_v\dfrac{\Delta s_k^{a}}{\Delta t_k}
        +(1-\lambda_v)\bar v_{k-1}$
        \State $K_k\leftarrow
        \lambda_K
        \operatorname{clip}\!\left(
        \dfrac{v_k}{\max(\bar v_k,\epsilon)},K_{\min},K_{\max}
        \right)
        +(1-\lambda_K)K_{k-1}$
    \Else
        \State $\mathbf{q}_k^d\leftarrow \mathbf{q}_{k-1}^d$, 
        $K_k\leftarrow K_{k-1}$, 
        $\bar v_k\leftarrow \bar v_{k-1}$
    \EndIf

\EndWhile
\end{algorithmic}
\end{algorithm}

\section{Experiments and Results}

The experiments were designed to evaluate the complete cybernetic performance of RL-MACRO across four critical dimensions: latent temperature perception, adaptive policy response, continuous spatial execution, and cross-sample robustness. To this end, two distinct experimental paradigms were utilized. First, line-milling experiments on bovine ribs---offering continuous IR measurement and a repeatable cortical--cancellous--cortical structural transition---were conducted to quantitatively evaluate temperature state observability and baseline adaptive regulation. Subsequently, realistic spiral craniotomy experiments on six \textit{ex vivo} goat skulls were performed to assess spatial action continuity and robust closed-loop adaptation under complex cranial geometries and severe tissue heterogeneity.

\subsection{Rib Line-Milling Experiment}

Fresh bovine ribs offer a repeatable cortical--cancellous--cortical transition \cite{27}, providing a natural environmental disturbance to evaluate the adaptive response of RL-MACRO. The line-milling setup permits continuous IR thermal tracking, serving as a ground-truth temperature reference for the multimodal observer. To rigorously isolate the benefits of closed-loop adaptation, RL-MACRO was benchmarked against a constant policy derived from its own temporal mean actions, which preserves equivalent average operating conditions but lacks state-dependent regulation. Three repeated paired trials across three distinct rib samples yielded $9$ paired experimental datasets.

\begin{figure}[htbp]
\centering
\includegraphics[width=\columnwidth]{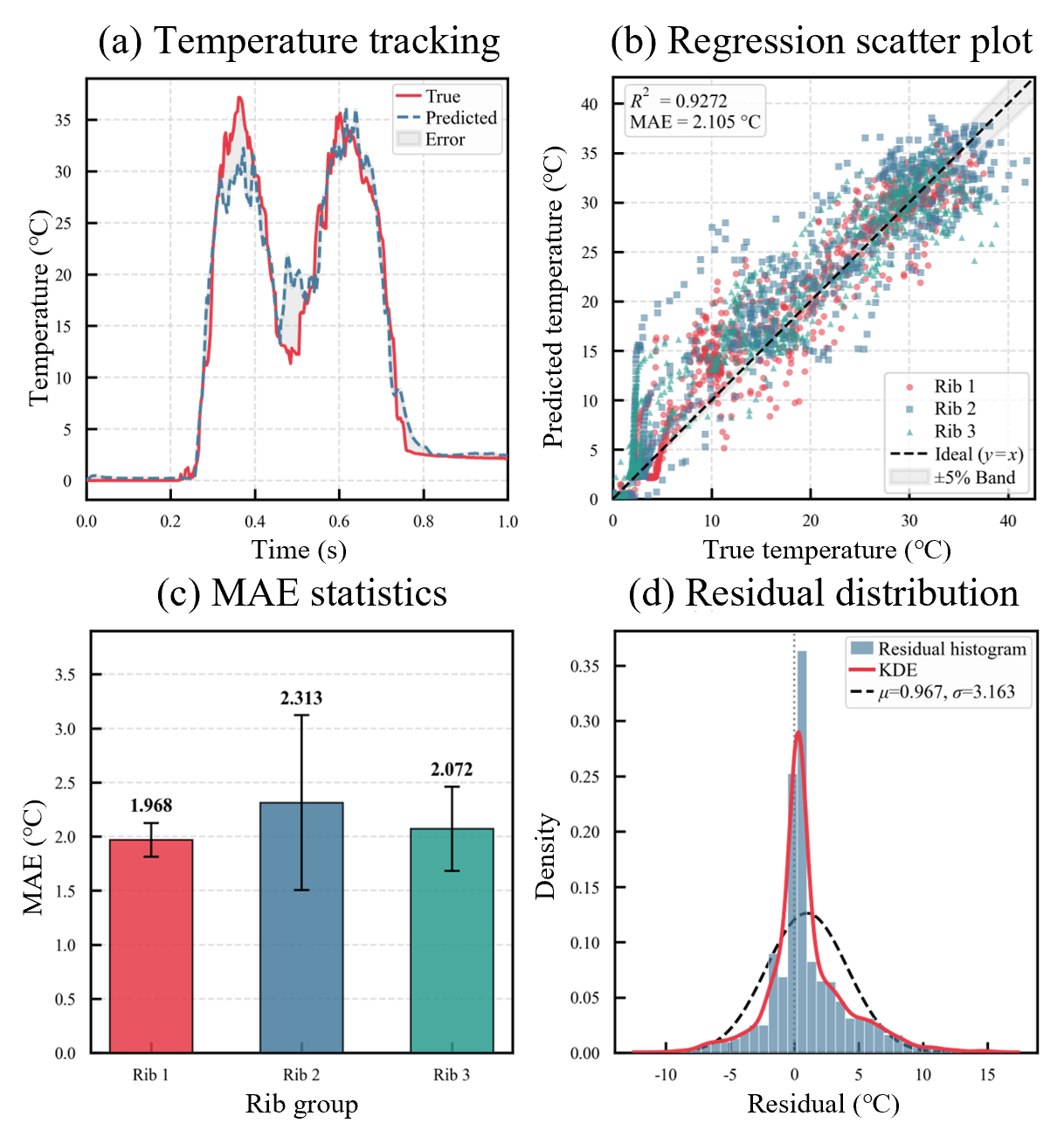}
\caption{Cross-sample online observability of the temperature state. (a) Temporal tracking across layered bone. (b) Regression between reconstructed and measured temperature rise. (c) MAE across three unseen rib samples. (d) Residual distribution.}
\label{fig:CNN-LSTM online}
\end{figure}

Fig.~\ref{fig:CNN-LSTM online} validates the online temperature observability. As shown in Fig.~\ref{fig:CNN-LSTM online}(a), the perceived temperature reliably tracks the measured thermal evolution across distinct bone layers. The absence of pronounced phase lag during rapid heating confirms that the causal sensory history encapsulates sufficient latent information. The cross-sample regression (Fig.~\ref{fig:CNN-LSTM online}(b)) achieves $R^2=0.9272$. Although slight dispersion exists---primarily due to the temporal mismatch between instantaneous contact mechanics and slower thermal inertia---the generalization across unseen specimens remains robust. Globally, the observer yields an MAE of $\SI{2.105}{\celsius}$ (Fig.~\ref{fig:CNN-LSTM online}(c)) with a minor positive bias of $\SI{0.967}{\celsius}$ (Fig.~\ref{fig:CNN-LSTM online}(d)), ensuring a conservative and reliable safety margin for the subsequent decision loop.

\begin{figure}[htbp]
\centering
\includegraphics[width=\columnwidth]{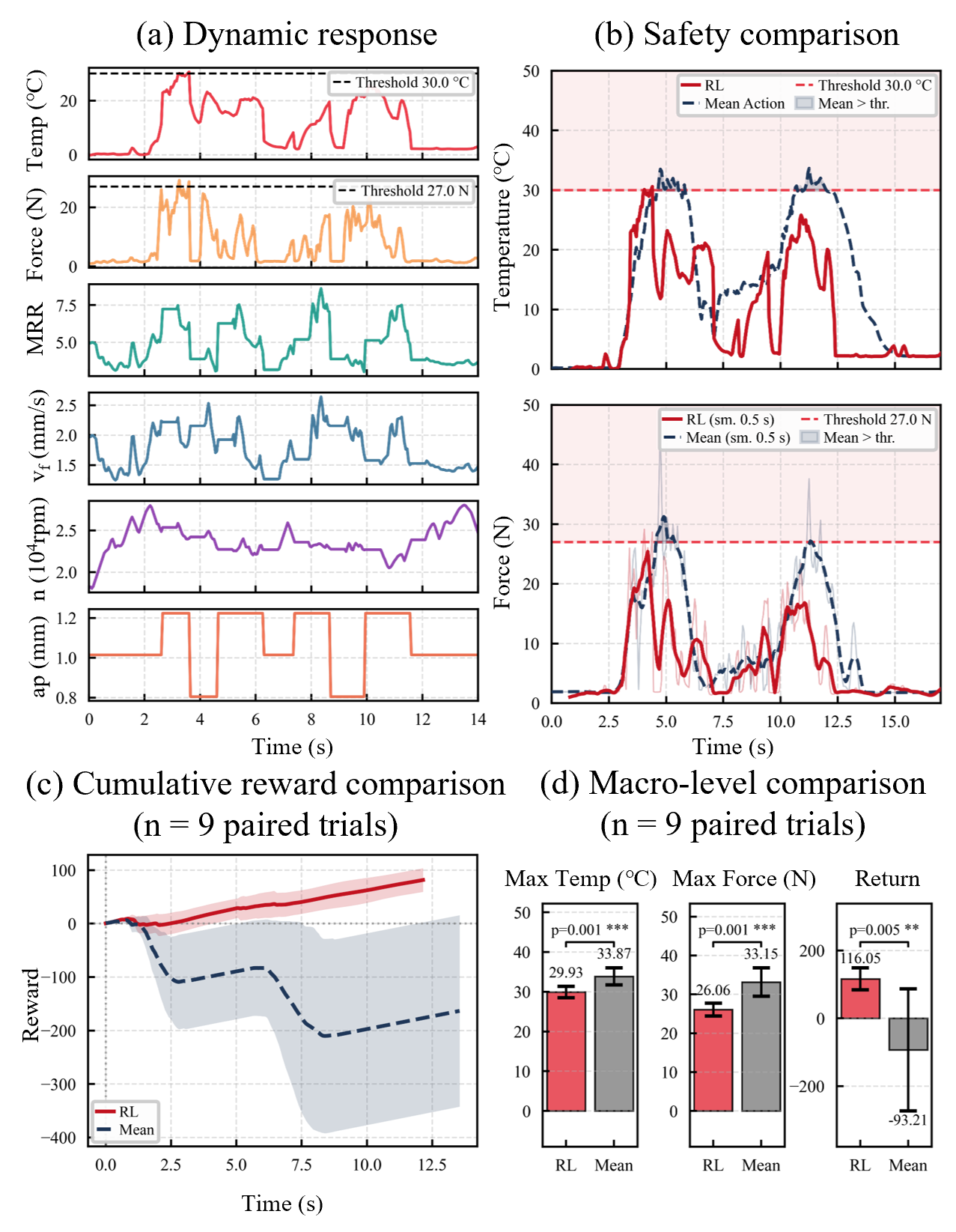}
\caption{Closed-loop adaptive behavior of RL-MACRO during rib milling. (a) Dynamic perception--decision--execution response. (b) Comparison of safety indicators. (c) Cumulative reward comparison ($n = 9$ paired trials). (d) Paired statistical comparison ($n = 9$ paired trials, $^{*}p<0.05$, $^{**}p<0.01$, $^{***}p<0.001$).}
\label{fig:RL online}
\end{figure}

Fig.~\ref{fig:RL online} illustrates the dynamic closed-loop response of the RL agent. In Fig.~\ref{fig:RL online}(a), variations in tool--tissue interaction continuously update the belief state, prompting immediate policy adaptation. The efficacy of the dual-head Actor architecture is explicitly verified: the macroscopic state classifier stabilizes the cutting depth ($a_p$) via smooth stepwise transitions, averting mechanical chattering, while the continuous head fine-tunes the feed rate ($v_f$) and spindle speed ($n$) at high frequency. Consequently, the system exhibits a robust deviation--adaptation--recovery pattern, decisively adjusting actions whenever the force or temperature approaches the predefined thresholds ($\SI{27.0}{\newton}$ and $\SI{30.0}{\celsius}$). 

Conversely, as depicted in Figs.~\ref{fig:RL online}(b) and (c), the non-adaptive baseline triggers severe safety penalty activations upon entering the hard cortical bone (gray shaded regions), causing the cumulative reward to plummet. Paired $t$-tests (Fig.~\ref{fig:RL online}(d)) confirm that RL-MACRO significantly mitigates maximum temperature and force excursions ($p=0.001^{***}$) while achieving an overwhelming advantage in episodic returns ($p=0.005^{**}$). These results compellingly demonstrate the agent's capability to navigate the Pareto-optimal frontier of safety and efficiency within uncertain biomechanical environments.

\subsection{Cross-Specimen Closed-Loop Validation on Ex Vivo Goat Skulls}

\begin{figure*}[t]
\centering
\includegraphics[width=\textwidth]{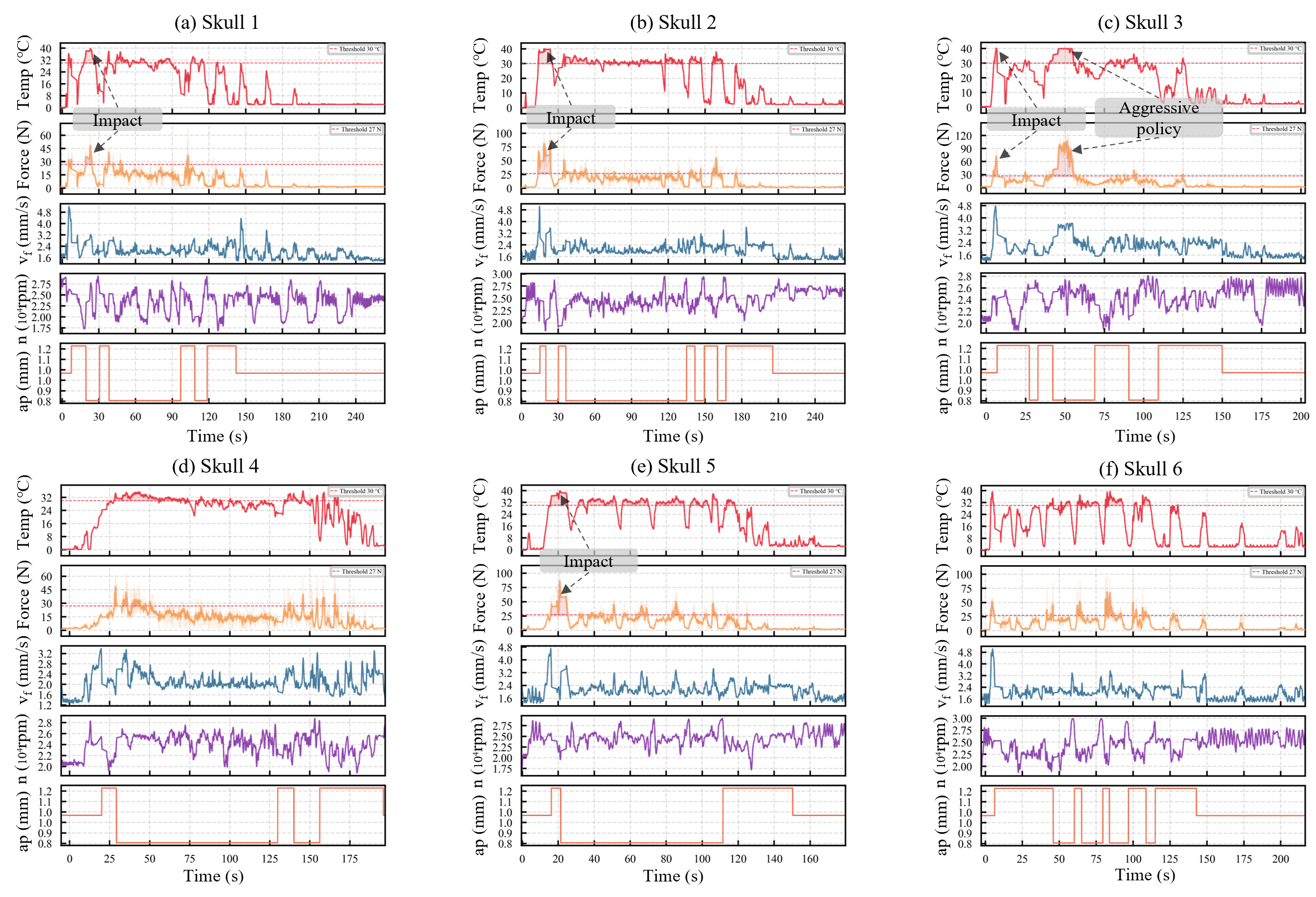}
\caption{Closed-loop adaptive responses during spiral craniotomy. (a)--(f) Continuous closed-loop control results across six isolated goat skulls from different individuals.}
\label{fig:skull rl}
\end{figure*}

\begin{figure*}[t]
\centering
\includegraphics[width=\textwidth]{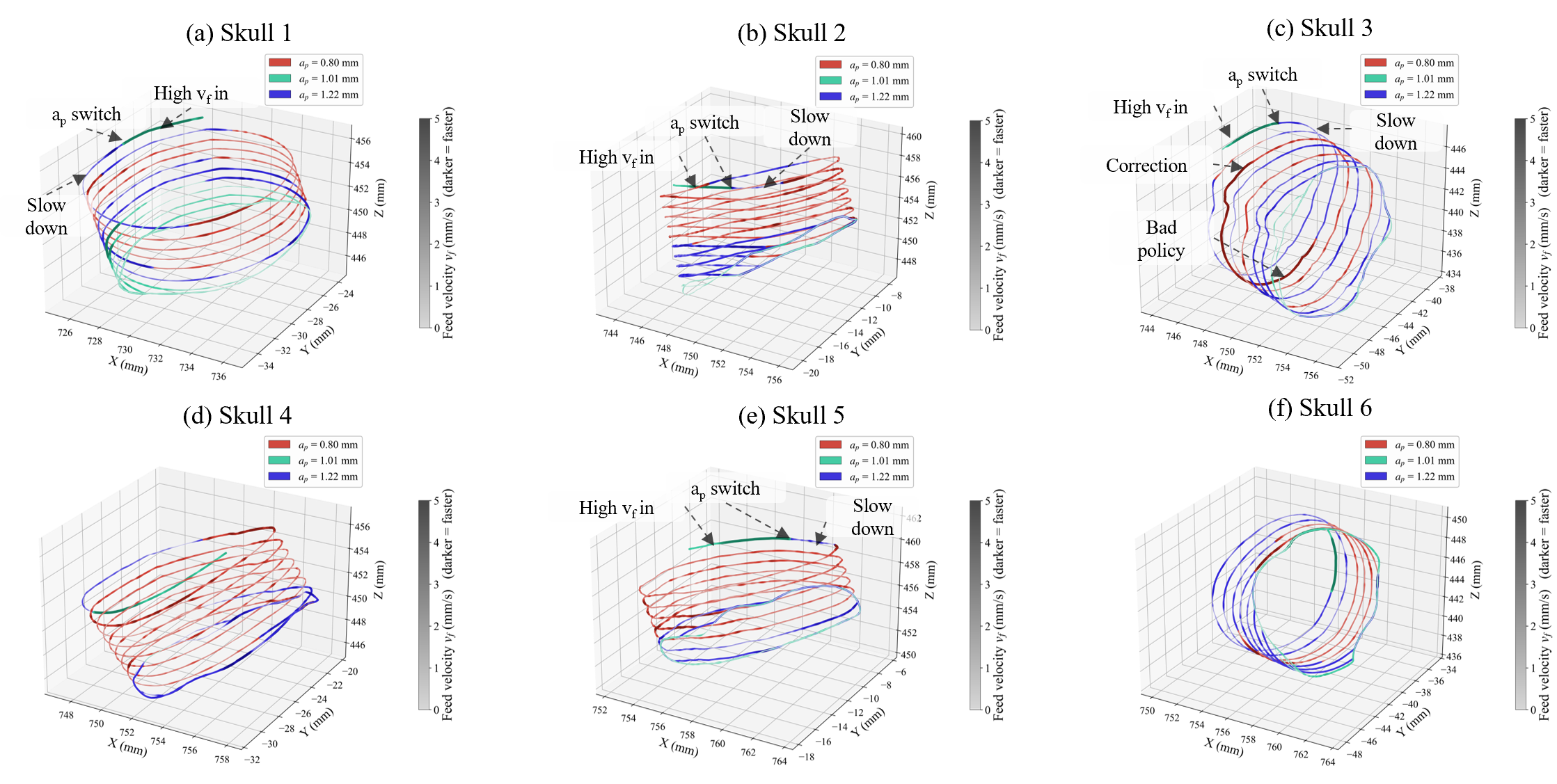}
\caption{Continuous action realization of adaptive policy. (a)--(f) Spatial execution pathways corresponding to the six isolated goat skulls.}
\label{fig:helix plots}
\end{figure*}

\begin{figure*}[t]
\centering
\includegraphics[width=\textwidth]{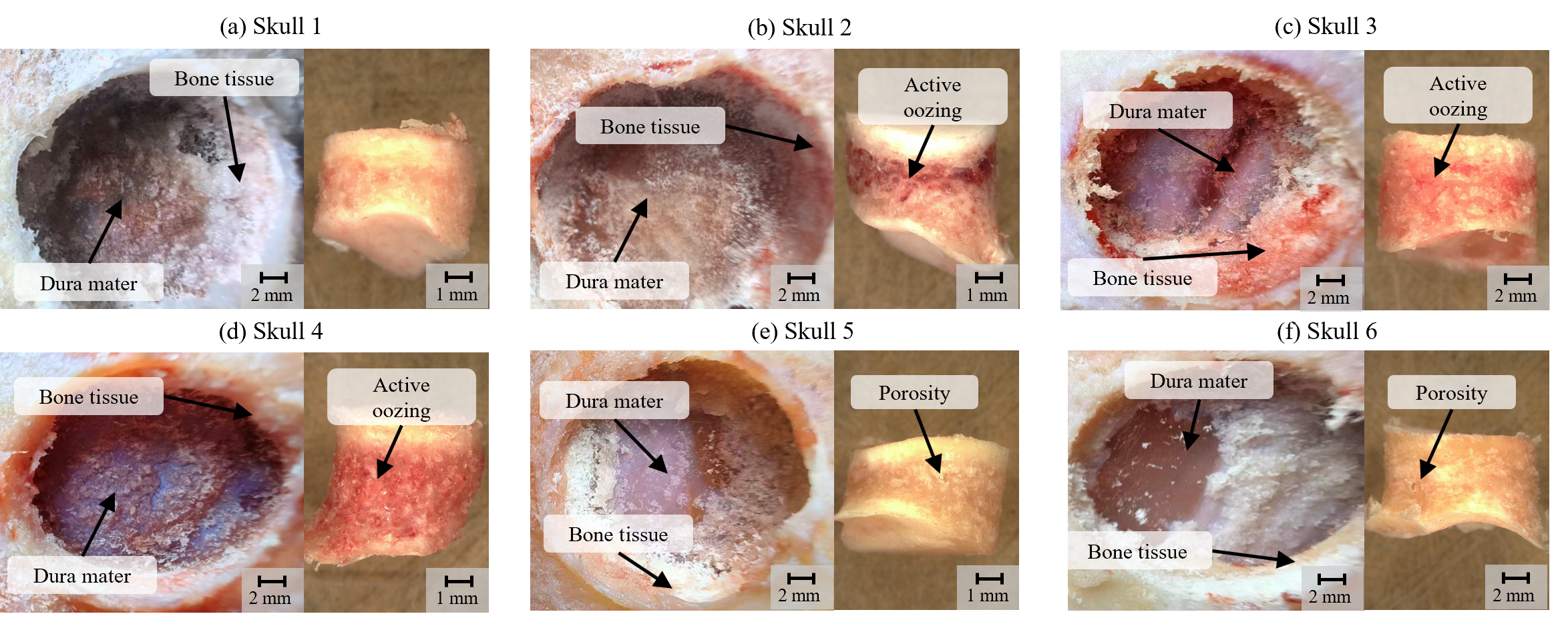}
\caption{Postoperative outcomes of the six \textit{ex vivo} craniotomy experiments. (a)--(f) Completely removed bone flaps and macroscopically intact cutting regions.}
\label{fig:real plots}
\end{figure*}

To evaluate the cybernetic loop's generalizability under complex anatomical geometries, spiral craniotomy was conducted on six \textit{ex vivo} goat skulls. These unseen samples present significant cross-specimen variations in curvature, thickness, and hardness. Due to severe physical occlusion during spiral milling, continuous IR monitoring is infeasible; hence, the aforementioned CNN--LSTM observer acts as the sole temperature state estimator to drive the online policy.

Fig.~\ref{fig:skull rl} illustrates the closed-loop temporal responses. Unlike the relatively uniform ribs, the profound spatial anisotropy of the skulls induced severe interaction challenges, with surgical durations varying from $\SI{210}{\second}$ to $\SI{260}{\second}$. Notably, during the initial penetration stage, the low tangential velocity at the cutter tip combined with the dense outer cortex frequently triggered force and temperature excursions (e.g., skulls a, b, c, and e). As these anomalies propagated into the belief state, RL-MACRO promptly orchestrated compensatory adjustments in feed rate and spindle speed, occasionally overriding the cutting depth via the implicit state classifier. The subsequent rapid decay of thermomechanical responses confirms a robust cybernetic loop: sensory deviation immediately induces policy compensation, whose physical efficacy is validated in the ensuing feedback cycle.

Furthermore, skulls (e) and (f) exhibit periodic force/temperature fluctuations likely induced by internal porosity. The agent seamlessly mirrors these disturbances with periodic parameter oscillations, proving its active engagement with the evolving environment rather than reverting to static conservatism. In skull (c), an aggressive feed-rate exploration at minimal depth around $\SI{40}{\second}$ caused a transient threshold breach, which the agent autonomously rectified by $\SI{60}{\second}$, highlighting its robust self-correcting capability against sudden physical extremes.

Fig.~\ref{fig:helix plots} visualizes the spatial realization of these discrete policy decisions. Variations in trajectory color and line intensity reflect dynamic updates to the cutting depth and feed rate, respectively. Driven by the dynamic re-planning module, all variable-pitch transitions maintain strict $C^0$ spatial continuity. The trajectories remain perfectly conformal to the irregular cranial surfaces without generating kinematic singularities or executing unphysical spatial jumps.

Fig.~\ref{fig:real plots} presents the postoperative outcomes. Complete bone-flap removal was achieved in all six samples. The cutting margins showed no visible carbonization, and no macroscopic dura mater damage was observed. These findings provide compelling preclinical evidence that the proposed perception--decision--execution framework can sustain safe, autonomous cutting across heterogeneous and previously unseen cranial samples. However, it must be acknowledged that macroscopic inspection alone does not constitute histological verification of thermal safety, and further \textit{in vivo} and microscopic validation remains necessary.

\section{Discussion}

This paper formulates autonomous robotic craniotomy as a partially observable cybernetic regulation problem, proposing the RL-MACRO framework to effectively close the ``perception--decision--execution'' loop. By reconstructing inaccessible interfacial temperatures via the CNN--LSTM multimodal observer, critical thermal states are elevated from post-process evaluations to active online feedback. In the decision layer, the implicit state classifier-guided dual-head IQL architecture achieves a superior safety--efficiency trade-off by coupling the macroscopic adaptation of cutting depth with high-frequency compensations of feed rate and spindle speed. Furthermore, the dynamic trajectory re-planning and velocity servo module seamlessly embodies these discrete policy outputs into $C^0$-continuous spatial motions. Validations on six \textit{ex vivo} goat skulls confirmed that this cybernetic loop successfully generalized to unseen, heterogeneous geometries without manual re-parameterization.

Despite its robust online performance, this data-driven offline RL methodology exhibits inherent limitations bound by the quality and state coverage of the training dataset. The current policy converged to safety thresholds of approximately $\SI{27}{\newton}$ (force) and $\SI{30}{\celsius}$ (temperature rise, $\Delta T$). In a clinical scenarios, the initial baseline temperature of bone exposed to the operating room environment is approximately $30^\circ\mathrm{C}$ \cite{28}, an unmitigated $\Delta T = \SI{30}{\celsius}$ risks thermal necrosis. However, considering that continuous saline irrigation typically attenuates the actual temperature rise to roughly $35\%$ of its uncooled value \cite{29}, the absolute tissue temperature is expected to remain between $\SI{40}{\celsius}$ and $\SI{55}{\celsius}$. While this aligns with the established thermal necrosis bounds ($\SI{47}{\celsius}$--$\SI{55}{\celsius}$) \cite{30}, inevitable intraoperative fluctuations in saline temperature and flow rates imply that this threshold remains marginally aggressive. The agent essentially converged to a sub-optimal extremum constrained by the available offline exploration boundaries. 

Moreover, although the IQL algorithm is inherently resilient to extrapolation errors \cite{25}, theoretical risks persist when confronting out-of-distribution (OOD) anatomical anomalies. Because our offline dataset was exclusively derived from bovine ribs, deploying the policy on goat skulls constituted a direct zero-shot domain transfer. This fundamental distribution shift elucidates the slightly amplified control fluctuations and the transiently aggressive policy excursion observed in the skull experiments (e.g., Fig.~\ref{fig:skull rl}(c)). 

Consequently, future efforts will proceed along two primary avenues. First, constructing a large-scale, high-diversity offline dataset that inherently encompasses cranial domain data alongside extreme machining scenarios. Second, developing a high-fidelity, thermodynamically coupled dynamic simulation model for human bone cutting, which will serve to pre-validate policy robustness and further guarantee the absolute safety of the RL agent prior to clinical deployment.

\section{Conclusion}

This paper presented RL-MACRO, a cybernetic closed-loop framework for adaptive robotic craniotomy under partial observability. Within this framework, a CNN--LSTM observer fuses force and sound feedback to reconstruct inaccessible interfacial temperatures. This latent state, alongside macroscopic cutting-regime probabilities, forms a robust belief state. A dual-head offline RL agent then maps this state into coordinated machining actions, which are seamlessly translated into spatially continuous robotic motions via dynamic trajectory re-planning and velocity servoing, thereby closing the perception--decision--execution loop.

Experimental validations on bovine ribs and six \textit{ex vivo} goat skulls confirmed robust temperature observability, superior thermomechanical regulation over constant-parameter baselines, and proactive recovery from transient physical excursions. The system successfully removed all bone flaps from irregular cranial surfaces without macroscopic thermal necrosis or dura mater damage. These findings establish a unified, data-driven cybernetic paradigm for safe autonomous osteotomy. Future work will focus on expanding cranial-domain offline datasets and advancing toward \textit{in vivo} histological validations.

\end{document}